\documentclass[journal]{IEEEtran}

\ifCLASSINFOpdf
\else
\fi

\IEEEoverridecommandlockouts

\usepackage[T1]{fontenc}
\usepackage{textcomp}

\usepackage{xcolor}
\usepackage{soul}
\usepackage{framed}

\colorlet{shadecolor}{yellow}

\definecolor{bgblue}{RGB}{60,90,120}
\definecolor{bglight}{RGB}{220,230,240}

\usepackage[cmex10]{amsmath}
\usepackage{amssymb}
\usepackage{mathtools}
\usepackage{bm}
\usepackage{physics}

\usepackage{array}
\usepackage{mdwmath}
\usepackage{mdwtab}
\usepackage{eqparbox}
\usepackage{breqn}

\usepackage{graphicx}
\graphicspath{{./images/}}

\usepackage{tikz}
\usetikzlibrary{arrows.meta,positioning}

\usepackage{multirow}
\usepackage{diagbox}
\usepackage{makecell}
\usepackage{threeparttable}
\usepackage{adjustbox}
\usepackage{rotating}

\usepackage[ruled,vlined]{algorithm2e}

\SetCommentSty{mycommfont}

\usepackage{pifont}
\usepackage{bbding}
\usepackage{wasysym}

\usepackage[
  separate-uncertainty = true,
  multi-part-units = repeat
]{siunitx}

\usepackage[nocompress,space]{cite}
\usepackage{url}
\usepackage{orcidlink}

\usepackage{hyperref}

\usepackage{cleveref}

\usepackage{pdfrender}

\usepackage{float}

\usepackage{blindtext}
\usepackage{lipsum}

\newif\ifshowmods
\showmodstrue     

\ifshowmods
                \newcommand{\Ared}[1]{\textcolor{red}{#1}}

\else
                \newcommand{\Ared}[1]{#1}

\fi

\title{

Glass-Box Deep Learning for 
FDIA
Detection in 
Nonlinear 
Automatic Generation Control:\\ A Kolmogorov–Arnold Network Approach

}

\author{
        
        Ahmad~Mohammad~Saber\orcidlink{0000-0003-3115-2384},~\IEEEmembership{Member,~IEEE,}
        Alok Paranjape\orcidlink{0009-0006-2766-5282},
        Jehad Jilan\orcidlink{0009-0008-1359-2515},
        Niranjana Naveen Nambiar\orcidlink{0009-0005-7031-9441},
        Amr~Youssef\orcidlink{0000-0002-4284-8646},~\IEEEmembership{Senior Member,~IEEE,}
        and~Deepa~Kundur\orcidlink{0000-0001-5999-1847},~\IEEEmembership{Fellow,~IEEE}
\thanks{Ahmad Mohammad Saber, Alok Paranjape, Jehad Jilan,  Niranjana Naveen Nambiar,  and Deepa Kundur are with 
the Department of Electrical and Computer Engineering, University of Toronto, Toronto, ON M5S 1A1, Canada
(e-mails: \href{mailto:ahmad.m.saber@ieee.org}{ahmad.m.saber@ieee.org};
\href{mailto:alok.paranjape@mail.utoronto.ca}{alok.paranjape@mail.utoronto.ca};
\href{mailto:jehad.jilan@mail.utoronto.ca}{jehad.jilan@mail.utoronto.ca};
\href{mailto:niranjana.nambiar@mail.utoronto.ca}{niranjana.nambiar@mail.utoronto.ca};
\href{mailto:dkundur@ece.utoronto.ca}{dkundur@ece.utoronto.ca})}
\thanks{Amr Youssef is with the Concordia Institute for Information Systems Engineering (CIISE), Concordia University, Montreal, QC, Canada
(e-mail: 
\href{mailto:youssef@ciise.concordia.ca}{youssef@ciise.concordia.ca}).}
}

\begin{document}

\maketitle



\begin{abstract}

Automatic Generation Control (AGC) plays a critical role in maintaining power balance across multi-area power systems. However, its complete reliance on remotely communicated measurements makes it susceptible to cyber-induced False Data Injection Attacks (FDIAs), which can alter measurement values and destabilize system operation. Unlike prior studies that employ black-box Deep Learning (DL) models for FDIA detection, this paper proposes an interpretable and accurate Kolmogorov–Arnold Network (KAN)-based framework for detecting FDIAs in AGC systems, explicitly accounting for nonlinearities often overlooked in previous work.
The proposed KAN model effectively identifies FDIAs in nonlinear AGC systems using only AGC measurements. Moreover, KANs inherently facilitate the extraction of symbolic equations, a capability absent in conventional DL models. After training, these equations can be directly utilized for FDIA detection, enhancing interpretability without compromising accuracy.
The framework is trained offline to learn the nonlinear relationships among AGC measurements under diverse normal operating conditions and under FDIA scenarios that manipulate those measurements. Following training, pruning, and fine-tuning, symbolic expressions describing the model’s decision logic are extracted. Both the trained KAN model and the extracted symbolic expressions are subsequently employed and evaluated for FDIA detection.
Our results using a benchmark power system demonstrate that the proposed KAN-based framework and its associated symbolic representations accurately detect FDIAs targeting nonlinear AGC systems while preserving AGC reliability. The approach outperforms existing methods and provides a robust, interpretable mechanism for verifying AGC measurement authenticity against potential FDIAs.


\end{abstract}

\begin{IEEEkeywords}
Automatic Generation Control, Kolmogorov-Arnold network, cyber-physical security, machine learning
\end{IEEEkeywords}

\section{Introduction}



\IEEEPARstart{A}{utomatic}
Generation Control (AGC) plays a central role in multi-area power systems by maintaining system frequency and regulating tie-line power flows to ensure continuous balance between generation and demand \cite{Musleh2023}. AGC continuously adjusts the generation output of participating units based on the remotely communicated frequency and tie-line power deviation measurements. However, these measurements are transmitted over open communication media using protocols, such as Distributed Network Protocol~3 (DNP3) or the Inter-Control Center Communications Protocol (ICCP)  \cite{ayad2022mitigation}, opening door for False Data Injection Attacks (FDIAs) against AGC measurements. Owing to its pivotal position in grid regulation, malicious entities can manipulate the AGC communicated measurements,  misleading the controller and potentially causing frequency excursions and instability issues \cite{He2020,Musleh2023,saber2022anomaly}. Thus, it is crucial to detect possible FDIAs against AGC measurements. 



FDIA detection in AGC systems has been increasingly studied, with the detection methods generally classified into traditional model-based and learning-based approaches. Model-based techniques, including residual thresholding and state estimation consistency checks, rely on linearized power system models to identify measurement inconsistencies \cite{Zhao2019,ayad2022mitigation}. Model-based require construction of precise equivalent models for the power system, which is cumbersome, and these techniques cannot precisely capture the AGC systems nonlinearities. 
%
With the rise of deep learning (DL), different DL-based methods have emerged as powerful tools for FDIA detection, being capable of learning the underlying complex relationships and nonlinearities in  the target component's measurements \cite{saber2025adversarial}.
Techniques such as Multi-Layer Perceptrons (MLPs), Long Short-Term Memory (LSTM) networks and Autoencoders have been proposed for FDIA detection in AGC systems \cite{Musleh2023,Abughali2025,ayad2022mitigation,shabar2025machine}. 
%
%
However, existing work on DL-based FDIA detection in AGC systems has the following limitations:

\begin{itemize}
    \item Existing deep learning models are opaque and lack interpretability. Their internal representations provide little insight into the reasoning behind classification outcomes, which hinders operator trust, limits forensic analysis, and poses challenges for deployment in safety-critical infrastructures.
    
    
    \item Most DL-based FDIA detection frameworks are trained and validated on datasets generated from linearized AGC representations that overlook GDB, GRC, and time-delay effects. Consequently, their learned mappings may not capture the full range of nonlinear responses encountered in real systems, leading to reduced robustness in practice.
    
\end{itemize}

Recently, there has been a growing interest in enhancing the interpretability of black-box DL-based models employed in critical purposes like FDIA detection and important infrastructure like power systems, thus enhancing the user/operator's trust in these models. A DL-based method with post-hoc gradient-weighted class activation mapping was proposed for enhancing interpretability of FDIA detection in power distribution system state estimation \cite{dehbozorgi2025false}. 
A  bidirectional long-short term memory method with post-hoc Shapley additive explanations was proposed for  FDIA detection in industrial internet of things systems \cite{attique2024explainable}.
Most recently, Kolmogorov–Arnold networks, a novel category of interpretable neural-networks, were proposed for short-term load forecasting in power systems\cite{jiang2024novel}, impedance identification of inverter-based resources \cite{shen2025kolmogorov}, electrical energy systems modeling \cite{zhou2025glass}, and hyperspectral change detection \cite{seydi2025kolmogorov}, offering intrinsically interpretable and accurate solutions.




This paper introduces an interpretable AI-based framework for detecting FDIAs in nonlinear AGC systems, leveraging Kolmogorov–Arnold Networks (KANs), a recently developed class of deep learning models inspired by the Kolmogorov–Arnold representation theorem \cite{kolmogrov_arnold_network_paper}.
Unlike conventional neural networks that rely on fixed activation functions, KANs employ learnable univariate functions along their connections, enabling them to approximate complex multivariate relationships in the AGC measurements, thereby enhancing FDIA detection, while retaining a clear functional structure. Further, through a symbolification process, trained KANs can produce human-readable symbolic expressions that represent the KAN model’s learned mapping between AGC features and detection decisions.
However, direct symbolic extraction from an unrefined KAN model often leads to loss of accuracy because the resulting expression may not perfectly approximate the locally fitted B-spline representations. To overcome this limitation, this paper employs a two-step refinement strategy involving network pruning followed by fine-tuning, which simplifies the model’s structure and restores its predictive fidelity. The outcome is an interpretable symbolic formula that closely matches the original model’s performance, achieving both transparency and accuracy in FDIA detection.
The key contribution of this paper can be summarized as follows:

\begin{itemize}
    \item 
    We propose a KAN-based approach for detecting FDIAs that manipulate the remotely communicated AGC measurements. The framework explicitly models AGC nonlinearities, ensuring realistic representation of operational dynamics under both normal and attack conditions.
    %
    The proposed KAN architecture enables the derivation of closed-form symbolic expressions that describe the learned detection logic. These expressions provide insight into how measurement deviations contribute to attack classification, improving operator transparency and trust. Further, the proposed framework can timely detect FDIAs.
    

    \item To demonstrate the above contribution, the proposed framework is validated under a comprehensive set of FDIAs of different types and trajectories, and FDIA-free disturbances, performed using a detailed two-area nonlinear AGC test system incorporating GDB, GRC, and communication delays nonlinearities. Our results confirm that the proposed KAN model achieves 97.28\% accuracy, and the extracted symbolic formula from the pruned and fine-tuned model achieves 95.9\% accuracy with an attack detection time of less than 0.16 seconds, demonstrating the feasibility of interpretable high-performance high-speed FDIA detection in AGC systems.
\end{itemize}


The remainder of this paper is organized as follows. Section~\ref{section:Preliminaries} presents the preliminaries, including AGC dynamics, nonlinear characteristics, and the threat model. Section~\ref{section:Methodology} describes the proposed methodology, covering KAN fundamentals, the pruning and fine-tuning process, and symbolic extraction. Section~\ref{section:Results} provides experimental results and interpretability analysis, followed by concluding remarks in Section~\ref{section:Conclusion}.

\section{Preliminaries and Threat Model}
\label{section:Preliminaries}

\begin{figure}[t!]
\centering
\includegraphics[width=1\linewidth]{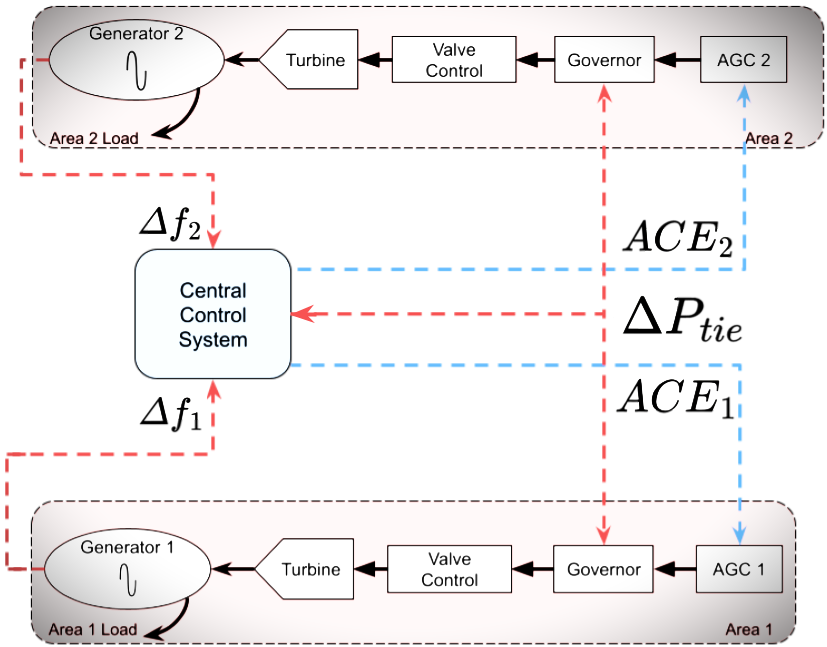}
\caption{Information flow diagram of a two-area power system with two generators equipped with AGC systems. AGC communicated measurements (\Ared{- -}) can be manipulated by cyber-attackers.}
\label{fig:information}
\end{figure}

AGC plays a central role in maintaining the balance between power generation and demand across interconnected control areas. It continuously monitors system frequency and tie-line power exchange, ensuring that deviations from nominal values are corrected through coordinated generator output adjustments. Any disturbance, whether a legitimate fluctuation such as a load change or generator outage, or a malicious cyber intrusion, can alter these measurements and, consequently, the control signal transmitted within the AGC loop \cite{shabar2025machine}. The basic information flow in a two-area AGC system and the nonlinearities inherent in its components are illustrated in 
Fig. \ref{fig:information}. 

The AGC calculates a corrective control signal known as the Area Control Error, which combines deviations in both frequency and tie-line power flow. It is expressed as

\begin{equation}
ACE = \Delta P_{tie} + B \, \Delta f
\end{equation}

\noindent where $\Delta P_{tie}$ is the deviation between scheduled and actual tie-line power exchange, $\Delta f$ is the system frequency deviation, and $B$ is the frequency bias factor for the control area \cite{ullah2021automatic}. The control objective is to drive the ACE toward zero by adjusting generator setpoints so that inter-area power flows remain within contractual limits and system frequency is restored to nominal values \cite{huang2018online}. The process operates continuously at sampling intervals of a few seconds, making it highly dependent on timely and accurate measurement data.



\begin{figure}[t!]
\centering
\includegraphics[width=1\linewidth]{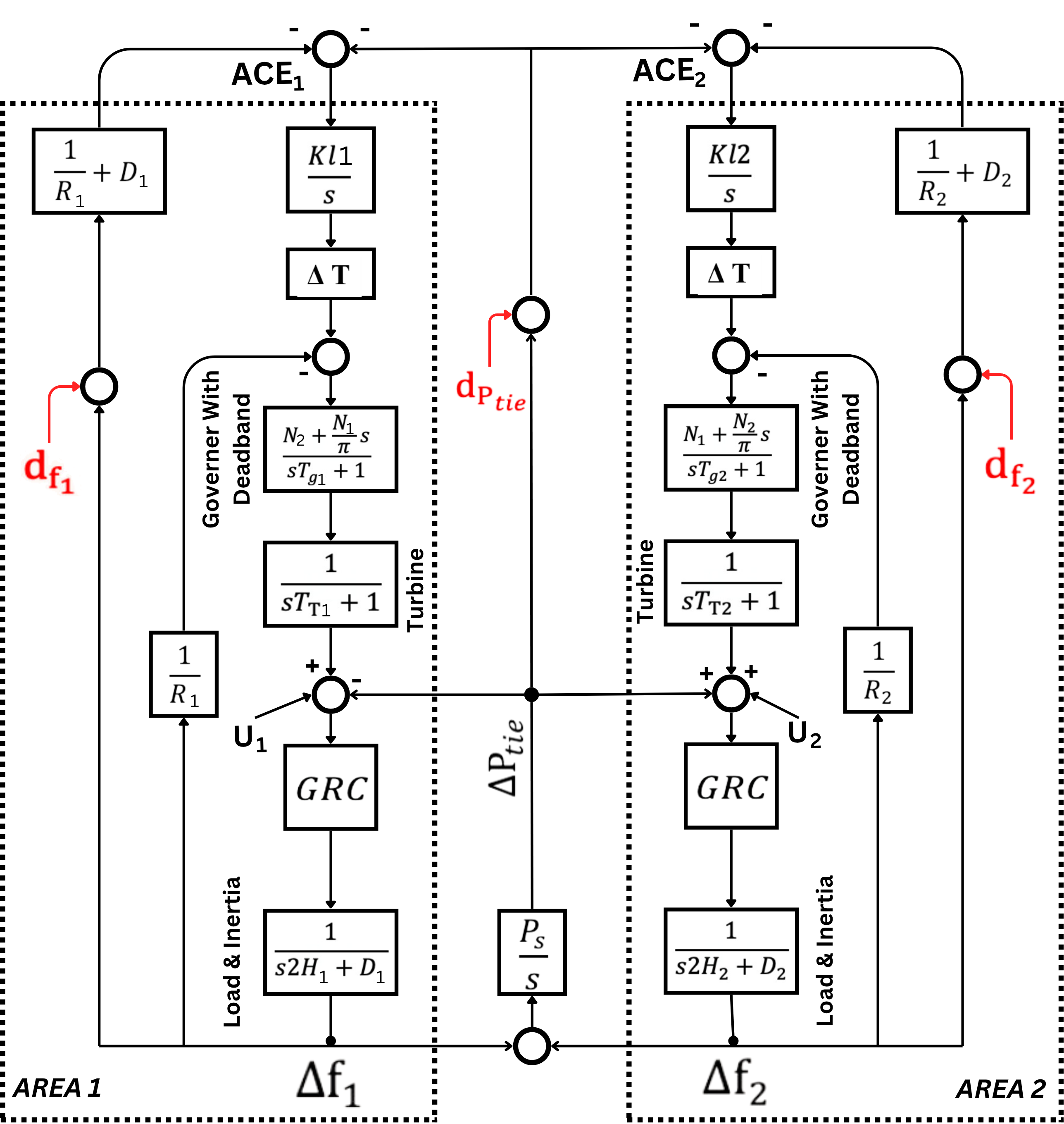}
\caption{ Block diagram  of a two-area AGC system including inherent AGC nonlinearities.}
\label{fig:nonlinearities}
\end{figure}

In practical operation, AGC dynamics are influenced by several nonlinearities that shape the system’s response to both legitimate disturbances and hence cannot be ignored  \cite{ayad2022mitigation,Musleh2023,shabar2025machine}. One such nonlinearity is the governor dead band (GDB), which introduces an insensitive region in the governor response,
to avoid excessive valve movement and mechanical wear \cite{khalaf2018joint}. Within this band, small frequency deviations do not trigger control action, resulting in a piecewise control characteristic that can mask subtle anomalies. Another significant nonlinearity is the generation rate constraint (GRC), which limits the rate of change of generator output \cite{khalaf2018joint,ayad2022mitigation}.
This restriction preserves turbine integrity but also slows the control response to rapid deviations, including those caused by cyber manipulations. Additionally, communication and actuation time delays, $\Delta \textbf{T}$, arise from both communication latency and physical actuation lags, creating temporal windows during which the control system operates on outdated information \cite{ayad2022mitigation}. These characteristics collectively make AGC a nonlinear, time-delayed feedback system in which small or gradual perturbations can propagate in non intuitive ways. Illustration of AGC with nonlinearities is shown in Fig. \ref{fig:nonlinearities}. Ignoring such nonlinearities  results in unrealistic system representations that fail to generalize to actual grid operation, a ma jor drawback of different related works, as discussed in \cite{ayad2022mitigation}.


As a cyber-physical system, the AGC relies on secure and reliable data exchange across geographically distributed components. Communication typically occurs through industrial protocols such as 
DNP3 or the  ICCP \cite{AGC_Class}. Despite improvements in protocol authentication and encryption, these links remain susceptible to FDIAs. A common class of attack exploiting this vulnerability is the FDIA, where an adversary deliberately alters measurement or control signals to mislead the controller while remaining undetected by conventional
detection schemes \cite{tan2016optimal}. FDIAs may be executed by compromising field sensors, programmable logic controllers, or intermediate communication nodes. Their design often relies on gradual manipulations, including ramp, step, or pulse functions, introduced in coordination across multiple measurement points. Such perturbations exploit the physical tolerances of the AGC system, causing deviations that remain within expected operational margins and are thus perceived as normal fluctuations \cite{shabar2025machine, tan2016optimal}. When observed at an individual time instant, these changes resemble measurement noise; yet, over time they can accumulate to produce significant deviations in ACE, potentially driving instability or frequency drift.

Detecting such attacks presents multiple challenges. First, the nonlinearities inherent in AGC dynamics complicate the separation between legitimate and malicious behavior, as similar measurement patterns can emerge from both. Second, the temporal nature of FDIAs requires models capable of capturing evolving dependencies across time. Third, given the safety-critical context of AGC operation, detection outcomes must be interpretable and traceable, allowing operators to understand the reasoning behind each alarm or classification. An effective detection framework must therefore integrate three essential capabilities: realistic modeling of AGC nonlinearities, temporal sensitivity to subtle and evolving attacks, and interpretability to support transparent decision-making.

The following section presents the proposed detection methodology, which employs an interpretable Kolmogorov–Arnold Network model to distinguish between normal and malicious AGC measurements while maintaining transparency and accuracy.

\section{Proposed Interpretable FDIA Detection Method for AGC Systems}
\label{section:Methodology}

This section presents the proposed interpretable learning framework for detecting FDIAs against nonlinear AGC systems.  
The framework leverages Kolmogorov–Arnold Networks to identify FDIAs that manipulate communicated AGC measurements, while maintaining model transparency through symbolic interpretability.

\subsection{Design Requirements and Challenges}

AGC systems are complex cyber–physical control loops that regulate system frequency and tie-line power by continuously exchanging measurement data between local areas and the control center.  
Their nonlinear response, influenced by factors such as governor deadband, generation rate constraint, and communication delay, complicates the modeling of their dynamic behavior.  
These nonlinearities not only affect control accuracy but also open potential vulnerabilities that adversaries can exploit to inject falsified data that mimic physical disturbances.

For an effective FDIA detection solution, three primary requirements must be satisfied:

\begin{itemize}
    \item The detection model must capture the nonlinear and time-varying behavior of AGC dynamics to accurately detect FDIAs and avoid misclassifying legitimate no-attack power system disturbances as attacks.
    \item The solution must rely solely on the AGC’s communicated operational measurements (frequency deviations and tie-line power) without requiring additional sensors or intrusive modifications, ensuring compatibility with existing infrastructure.
    \item Because AGCs operate within safety-critical infrastructure, operators must be able to understand why a model detects an FDIA. Therefore, the FDIA detection logic should be explainable in explicit mathematical terms, enabling inspection and validation against physical principles.
\end{itemize}

Meeting these requirements poses several design challenges.  
First, the high dimensionality of raw AGC time-series data makes it computationally demanding to apply deep models directly while maintaining interpretability.  
Second, conventional neural networks (e.g., MLPs or LSTMs) achieve high accuracy but act as black boxes, preventing operator trust and regulatory acceptance.  
Finally, the symbolic expressions extracted from interpretable models often lose accuracy relative to the trained model unless structural optimization (e.g., pruning and fine-tuning) is applied.  
The proposed framework addresses these challenges by combining a KAN architecture with feature-based input representation, structured pruning, and symbolic regression to achieve a balance between interpretability and accuracy.

\subsection{Kolmogorov–Arnold Network Formulation}

KANs are grounded in the Kolmogorov–Arnold representation theorem, which states that any multivariate continuous function \(F:[0,1]^n \!\rightarrow\! \mathbb{R}\) can be decomposed as
\begin{equation}
F(\mathbf{x}) = \sum_{k=1}^{2n+1}
\Phi_k\!\left( \sum_{l=1}^{n} \Psi_{k,l}(x_l) \right)
\label{eq:kan_theorem}
\end{equation}

\noindent where \(\Psi_{k,l}(\cdot)\) and \(\Phi_k(\cdot)\) are continuous univariate functions.  
Modern KANs extend this concept to deeper architectures with arbitrary width and depth, allowing  modeling of nonlinear mappings.  
For an input feature vector \(\mathbf{x}\!\in\!\mathbb{R}^n\), the output of one KAN layer is given by
\begin{equation}
\mathbf{h} = \Omega_{\mathrm{out}}\!\big(\Omega_{\mathrm{in}}(\mathbf{x})\big)
\end{equation}
where
\begin{equation}
\Omega_{\mathrm{in}}(\mathbf{x}) = 
\begin{bmatrix}
\Psi_{1,1}(x_1) & \cdots & \Psi_{1,n}(x_n) \\
\vdots & \ddots & \vdots \\
\Psi_{2n+1,1}(x_1) & \cdots & \Psi_{2n+1,n}(x_n)
\end{bmatrix}
\end{equation}

\begin{equation}
\Omega_{\mathrm{out}} = [\,\Phi_1,\Phi_2,\ldots,\Phi_{2n+1}\,]
\end{equation}

\begin{figure}[t!]
    \centering
    \includegraphics[width=0.95\linewidth]{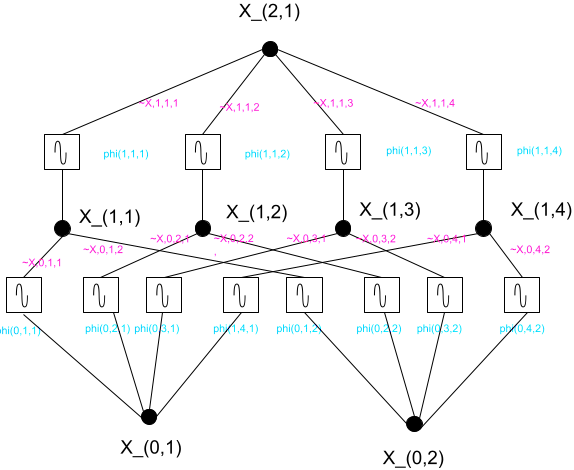}
    \caption{Illustration of KANs idea with learnable activation functions one edges.}
    \label{fig:kan_diagram_new}
\end{figure}

\noindent Stacking \(T\) such layers yields the overall mapping
\begin{equation}
\mathcal{F}_{\theta}(\mathbf{x}) =
\Omega_{T-1} \!\circ\! \Omega_{T-2} \!\circ\! \cdots \!\circ\! \Omega_{0}(\mathbf{x}),
\label{eq:kan_multilayer}
\end{equation}
where \(\theta\) denotes all trainable parameters.  
Each univariate edge function \(\Psi_{k,l}(\cdot)\) is parameterized as a B-spline expansion:
\begin{equation}
\Psi_{k,l}(x) = \sum_{p=1}^{P} c_{k,l,p}\, B_{p}(x)
\label{eq:bspline}
\end{equation}
with \(B_{p}(x)\) denoting the \(p\)-th spline basis and \(c_{k,l,p}\) the learned coefficients.  
As illustrated in  Fig. \ref{fig:kan_diagram_new}, 
KANs leverage  univariate functions on their edges that can be learned during KAN training, unlike traditional neural networks that have fixed activation functions.
This representation enables the model to approximate nonlinear dependencies in AGC measurements with high precision while maintaining a functional form amenable to symbolic interpretation.

The KAN parameters $\theta$ are optimized using the Binary Cross-Entropy with Logits Loss (BCEWithLogitsLoss), which combines a sigmoid activation with the binary cross-entropy objective in a numerically stable formulation.  
This loss directly operates on the raw output $\tilde{f}_{\theta}(\mathbf{x})$ (logit) of the model rather than its post-activation probability, ensuring stable gradient behavior during optimization.  
The training process employs the limited-memory Broyden–Fletcher–Goldfarb–Shanno (LBFGS) optimizer, which is well suited for smooth spline-based functions \cite{saputro2017limited}.  
Model convergence is monitored using validation accuracy and loss to prevent overfitting.

\subsection{Model Pruning and Fine-Tuning}



To improve interpretability and reduce model complexity, the trained KAN can be simplified through a pruning process that removes connections with negligible contributions. Specifically, edges whose influence magnitudes fall below a predefined threshold~$\tau$ are pruned according to

\begin{equation}
|\omega_{ij}| < \tau
\label{eq:pruning}
\end{equation}

\noindent where~$\omega_{ij}$ denotes the effective edge weight or influence magnitude, and~$\tau$ represents the pruning threshold.
After pruning, the model can be fine-tuned on the original training dataset using the aforementioned loss function  to restore accuracy while maintaining sparsity.
This process yields a more compact and interpretable analytical representation~$\tilde{f}(\mathbf{x})$ that preserves high fidelity to the unpruned model’s predictions.

\subsection{Symbolification and Interpretability}

After training, the learned numerical mapping \(\mathcal{F}_{\theta}(\mathbf{x})\) is approximated by a symbolic expression \(\tilde{f}(\mathbf{x})\) through a process termed \emph{symbolification}:
\begin{equation}
\tilde{f}(\mathbf{x}) = \sum_{i=1}^{M} \alpha_i f_i(\mathbf{x})
\label{eq:symbolic_form}
\end{equation}
where \(f_i(\mathbf{x})\) are analytic basis functions selected from a predefined set of functions, e.g., polynomial, exponential, logarithmic, trigonometric \cite{kan_repo}, and \(\alpha_i\) are fitted coefficients.  
%

\begin{figure*}[t!]
\centering
\includegraphics[width=0.95\linewidth]{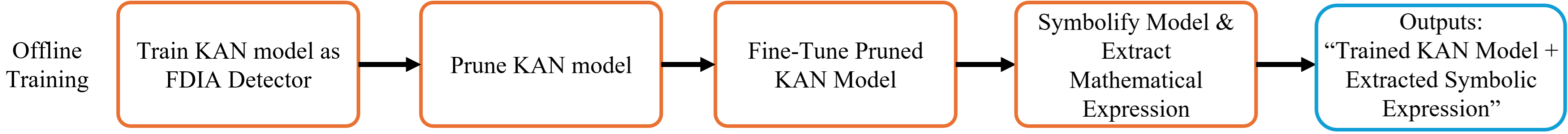}
\caption{Offline Training and Symbolification Workflow of the Proposed KAN-Based FDIA Detection Framework}
\label{fig:KAN_offline}
\end{figure*}

%
As the KAN model is optimized during training using the BCEWithLogitsLoss, where the model output $\tilde{f}(\mathbf{x})$ represents a raw logit rather than a post-activation probability. Accordingly, during inference and symbolic evaluation, the decision rule is defined directly on this logit output as
\begin{equation}
\hat{y} =
\begin{cases}
1  & \text{if } \tilde{f}(\mathbf{x}) > 0 \\
0  & \text{otherwise}
\end{cases}
\label{eq:logit_threshold}
\end{equation}
where \(\hat{y}=1\) indicates an FDIA, and \(\hat{y}=0\) corresponds to a normal operating condition.  
This thresholding approach is equivalent to applying a sigmoid activation followed by a $0.5$ cutoff, but avoids the redundant computation of $\sigma(z)=1/(1+e^{-z})$ during testing. Thus, a positive logit indicates a detected FDIA, while a non-positive logit corresponds to a normal operating condition.
This analytic form provides human-readable decision logic, allowing operators to identify dominant features and understand the rationale for each detection.
A summary of the of the proposed KAN framework's offline training process is depicted in Fig. \ref{fig:KAN_offline}. 


\subsection{Overall FDIA Detection Workflow}

\begin{figure*}[t!]
\centering
\includegraphics[width=0.85\linewidth]{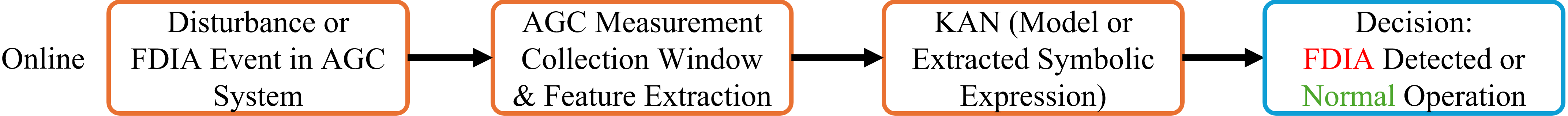}
\caption{Online operation.}
\label{fig:KAN_online}
\end{figure*}

When an AGC controller receives a new set of communicated measurements corresponding to a disturbance, which could be also an FDIA, the proposed model operates on a short temporal window of AGC measurements before and after the event.  
Let \(\mathbf{z}\) denote the multi-dimensional sequence of raw AGC measurements.
In other words, 
\(\mathbf{z}\) contains multiple AGC measurement channels, including measurements for tie-line power deviation \((\Delta P_{\text{tie}})\) and area frequency deviations \((\Delta f_1, \Delta f_2)\). Each measurement in \(\mathbf{z}\) is recorded as a time series.
%
Applying KANs directly to 
\(\mathbf{z}\)
would significantly increase dimensionality and reduce interpretability.  
Therefore, a feature extraction stage is leveraged to summarize each AGC measurement sequence into descriptive indicators that retain essential power system dynamics.  
From each measurement in \(\mathbf{z}\), e.g., $\Delta P_{tie}$, a set of features are computed, producing a feature vector \(\mathbf{x}\!\in\!\mathbb{R}^n\).
The KAN then operates on this reduced-dimension representation to distinguish measurements that correspond to genuine physical changes from measurements merely manipulated by deliberate false-data injections.
This representation reduces computational complexity and enables efficient symbolic regression without compromising the ability to distinguish normal and attacked conditions.
This paper employs a set of standard statistical metrics commonly used in FDIA detection research \cite{shefaei2021revealing}, including mean, standard deviation, minimum, maximum, skewness, and kurtosis, noting that proposed framework's primary focus is demonstrating the effectiveness of KANs. 
The proposed KAN-based framework flexibility also allows for the incorporation of alternative feature sets without modification.
%
%
Fig. \ref{fig:KAN_online} illustrates the online information flow diagram of the proposed KAN-based framework.
%
The proposed methodology relies solely on the communicated AGC measurements \(\mathbf{z}\), and integrates feature-based input representation, spline-based functional learning, pruning, and symbolic regression into a unified interpretable framework for FDIA detection in AGC systems considering their nonlinearities.  


\section{Performance Evaluation of the Proposed KAN-Based Framework }
\label{section:Results}



\subsection{Comprehensive FDIA Scenarios and Test System Description}

 The performance of the proposed approach is evaluated using the  Kundur two-area AGC benchmark system widely adopted in prior research \cite{ayad2022mitigation, Musleh2023, Abughali2025}, considering the system nonlinearities. The system models a two-area power system, where each area has of two 20 kV, 900 MVA generation units; one of which is equipped with an AGC system. 
 System parameters are summarized in Table \ref{tab:system_params}.
 This test system model also explicitly incorporates AGC nonlinearities GDB, GRC, and communication and actuation delays, which fundamentally affect the AGC’s dynamic response to both legitimate disturbances and malicious measurement perturbations. Their inclusion ensures that the generated data reflect the temporal complexity and saturation effects of real-world power system operation, making the dataset a rigorous test environment.  
 In the simulation setup, delay is modeled using a fixed delay incorporated into the AGC feedback loop, where the ACE at time \( k \) is computed using delayed measurements of frequency deviation \( \Delta f_i[k-d] \) and tie-line power deviation \( \Delta P_{\text{tie}}[k-d] \), with \( d \) denoting the number of discrete delay steps. This formulation captures the telemetry and processing latency inherent in real AGC communication channels. The GRC is modeled by limiting the rate of change in mechanical power output as \( |\Delta P_{m,i}[k] - \Delta P_{m,i}[k-1]| \leq \delta_{\max} \), ensuring physically consistent turbine behavior. Similarly, the GDB is represented as a dead zone in which no control action is taken when \( |\Delta f_i| < \Delta f_{\text{th}} \), reflecting industrial governor practices that suppress minor frequency deviations to reduce actuator wear \cite{khalaf2018joint,Abughali2025}.

 \begin{table}[!t]
\caption{Utilized Two-Area System Parameters\label{tab:system_params}}
\centering
\footnotesize
\begin{tabular}{c c l}
\Xhline{3\arrayrulewidth}
\rule{0pt}{2ex}  \textbf{Parameter} & \textbf{Value} & \textbf{Description} \\ 
\rule{0pt}{3ex}  $H_1, H_2$          & 5, 4        & Inertia constants (s)               \\ 
\rule{0pt}{3ex}  $D_1, D_2$          & 0.6, 0.3    & Damping coefficients (pu/Hz)        \\ 
\rule{0pt}{3ex}  $B_1, B_2$          & 20.6, 16.3  & Frequency bias (pu/Hz)              \\ 
\rule{0pt}{3ex}  $T_{g,1}, T_{g,2}$  & 0.2, 0.3    & Governor time constants (s)         \\ 
\rule{0pt}{3ex}  $T_{t,1}, T_{t,2}$  & 0.5, 0.6    & Turbine time constants (s)          \\ 
\rule{0pt}{3ex}  $R_1, R_2$          & 0.05, 0.0625 & Speed regulation (pu)              \\ 
\rule{0pt}{3ex}  $T_{12}$            & 2           & Synchronizing coefficient (pu)      \\ 
\rule{0pt}{3ex}  $\text{GDB}$        & 0.06        & Governor deadband (\%)             \\ 
\rule{0pt}{3ex}  $GRC_1, GRC_2$      & $\pm3$      & Gen.\ rate constraints (pu/min)     \\ 
\rule{0pt}{3ex}  $K_{\text{i},1},K_{\text{i},2}$ & 0.3, 0.3 & AGC integral gains (-)  \\ 
\Xhline{3\arrayrulewidth}
\end{tabular}
\end{table}

\begin{figure}[t!]
    \centering
    \includegraphics[width=0.9\linewidth]{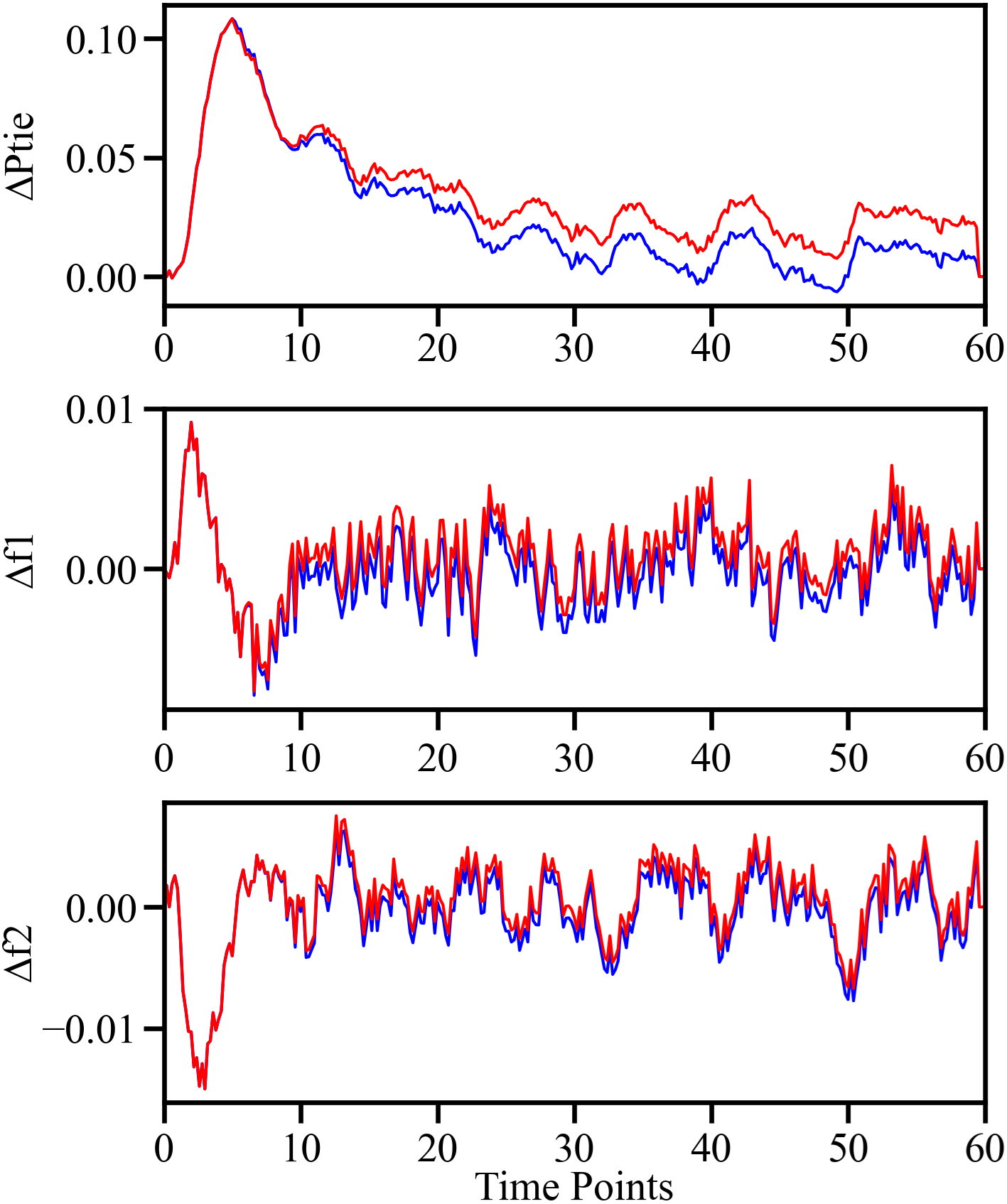}
    \caption{Representative AGC measurements under a normal (blue) disturbance and an FDIA (red).}
    \label{fig:timeline}
\end{figure}

A comprehensive dataset of FDIAs and normal disturbances is generated using this test system.
%
FDIAs are simulated as coordinated manipulations of one or more of the AGC measurements ($\Delta P_{\text{tie}}, \Delta f_1$, and $\Delta f_2$), with various attack attack types including ramp, step, and combined attacks of various magnitudes, introducing variability and complexity into the dataset.
%
Two examples of an FDIA and a disturbance without an FDIA from the generated dataset are shown in Fig. \ref{fig:timeline}.
Each generated sample covers 60 seconds, with a sampling interval of 0.2 seconds. Disturbance magnitudes are drawn from a normal distribution with zero mean and a standard deviation of 0.02 p.u., and attacks are triggered following disturbances to emulate realistic interference scenarios. Noise is incorporated into the simulations to reflect real-world uncertainties. Process noise and measurement noise are modeled as white Gaussian noise with zero mean and a standard deviation of $10^{-6}$ \cite{Musleh2023,ayad2022mitigation}. Process noise represents stochastic variations in system dynamics, while measurement noise captures sensor inaccuracies. To ensure the dataset includes a broad range of attack difficulties, an ACE limit is imposed to constrain post-attack deviations. Attacks exceeding this limit, set at 1.0, are rescaled to maintain a challenging detection environment, following \cite{Musleh2023,ayad2022mitigation}.
For each sample, the three communicated AGC measurements are recorded: tie-line power deviation ($\Delta P_{\text{tie}}$) and frequency deviations in Area~1 and Area~2 ($\Delta f_1$, $\Delta f_2$). These signals represent the only operational technology (OT) measurements transmitted between control areas and the AGC controller, reflecting realistic data availability for security monitoring.


For each scenario, the statistical descriptor features explained in Section \ref{section:Methodology}
are determined for each of the three AGC measurement sequences, yielding 18 features per scenario, as shown in Table~\ref{tab:variables}.  
\begin{table}[t!]
\centering
\caption{Extracted features and corresponding variables in \(\tilde{f}(\mathbf{x})\)}
\begin{tabular}{c l|c l}
\Xhline{3\arrayrulewidth}
\rule{0pt}{2ex} \textbf{Variable} & \textbf{Feature} & \textbf{Variable} & \textbf{ Feature} \\
\rule{0pt}{3ex}  $x_{1}$ & mean of $\Delta P_{\text{tie}}$   & $x_{10}$ & max of $\Delta f_1$ \\ 
\rule{0pt}{3ex}  $x_{2}$ & std of $\Delta P_{\text{tie}}$    & $x_{11}$ & skew of $\Delta f_1$  \\
\rule{0pt}{3ex}  $x_{3}$ & min of $\Delta P_{\text{tie}}$    & $x_{12}$ & kurtosis of $\Delta f_1$  \\
\rule{0pt}{3ex}  $x_{4}$ & max of $\Delta P_{\text{tie}}$    & $x_{13}$ & mean of $\Delta f_2$  \\
\rule{0pt}{3ex}  $x_{5}$ & skew of $\Delta P_{\text{tie}}$   & $x_{14}$ & std of $\Delta f_2$ \\
\rule{0pt}{3ex}  $x_{6}$ & kurtosis of $\Delta P_{\text{tie}}$ & $x_{15}$ & min of $\Delta f_2$ \\
\rule{0pt}{3ex}  $x_{7}$ & mean of $\Delta f_1$       & $x_{16}$ & max of $\Delta f_2$ \\
\rule{0pt}{3ex}  $x_{8}$ & std of $\Delta f_1$        & $x_{17}$ & skew of $\Delta f_2$ \\
\rule{0pt}{3ex}  $x_{9}$ & min of $\Delta f_1$        & $x_{18}$ & kurtosis of $\Delta f_2$ \\
\Xhline{3\arrayrulewidth}
\end{tabular}
\label{tab:variables}
\end{table}
These features capture both steady-state and transient characteristics of AGC responses surrounding each disturbance. The dataset contains 20,000 labeled instances evenly split between normal and attacked scenarios, providing a balanced and statistically diverse benchmark for model training and evaluation. 

The above dataset is randomly divided into 60\% training, 20\% validation, and 20\% testing the proposed method.

\subsection{Performance Evaluation Metrics}

To ensure a holistic evaluation and capture both AGC cybersecurity and dependability perspectives, the performance of the proposed method is quantified using a variety of standard metrics:
\begin{equation}
\text{Accuracy} = \frac{TP + TN}{TP + TN + FP + FN}
\end{equation}

\begin{equation}
\text{Recall} = \frac{TP}{TP + FN}
\end{equation}

\begin{equation}
\text{Precision} = \frac{TP}{TP + FP}
\end{equation}

\begin{equation}
\text{F1-score} = 2 \times \frac{\text{Precision}\times\text{Recall}}{\text{Precision}+\text{Recall}}
\end{equation}
where $TP$, $TN$, $FP$, and $FN$ denote true positives (detected FDIAs), true negatives (correctly classified non-attack instances), false positives, and false negatives, respectively. These indicators reflect the balance between sensitivity to attacks and resilience against false alarms, both essential for dependable AGC operation.

\subsection{Performance Evaluation Experiments, Model Training and  Settings}

Two experiments are conducted. \textit{Experiment~1} trains the KAN without pruning, serving as a performance benchmark. \textit{Experiment~2} applies pruning followed by fine-tuning to enhance interpretability and simplify the symbolic representation while maintaining accuracy.



\begin{figure}[t!]
    \centering
    \includegraphics[width=0.95\linewidth]{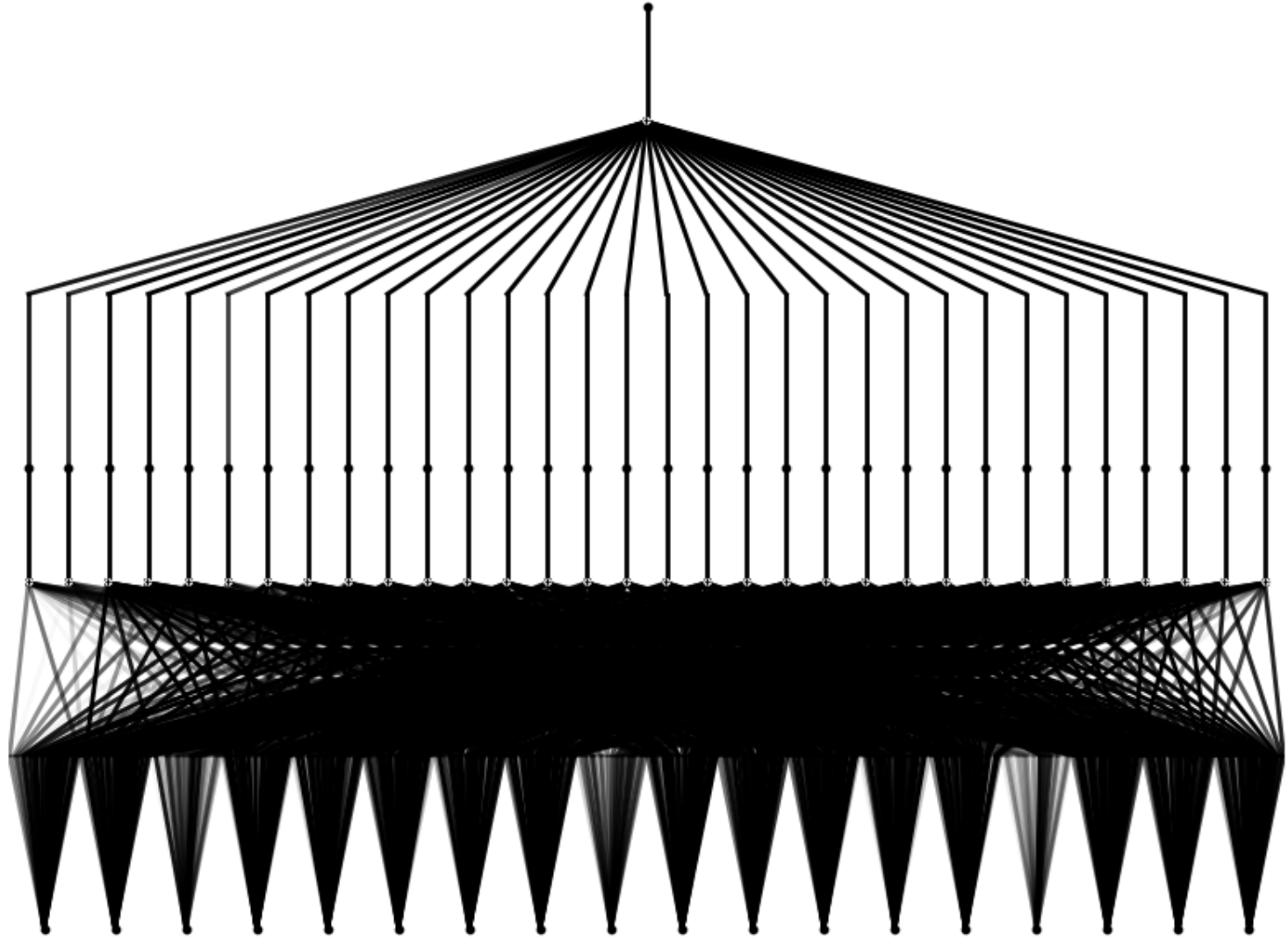}
    \caption{Illustration of the utilized KAN.}
    \label{fig:KAN_initial}
\end{figure}

In each experiment, a Kolmogorov--Arnold Network (KAN) model is trained for binary classification, distinguishing between normal and attacked AGC states. The model follows an input--hidden--output configuration with layer widths of $[n_{\text{in}},\, 32,\, 1]$, following \cite{kan_repo, kolmogrov_arnold_network_paper}, where $n_{\text{in}}$ denotes the input feature dimension. Each layer employs spline-based activation functions parameterized by a grid size of $3$ and a spline order of $k=3$. The base nonlinearity is the SiLU function, used for initializing spline coefficients. The model implementation (\texttt{MultKAN}) contains both numerical KAN layers and their symbolic counterparts: the numerical layers (\texttt{KANLayer}) are trained using B-spline activations, while the symbolic layers (\texttt{Symbolic\_KANLayer}) store analytic expressions extracted post-training through the symbolification process. This dual structure enables the same trained model to provide both high-accuracy numerical inference and interpretable symbolic formulas.
An illustration of the utilized KAN with above specifications is shown in Fig. \ref{fig:KAN_initial}.
 The LBFGS optimizer is used with a binary cross-entropy loss, and training proceeds for 50 epochs.  
The model is trained for $50$ optimization steps with a regularization coefficient of $\lambda = 0.001$ and a mini-batch size of $128$. 
Model pruning  is performed using the default value of ~$\tau = 10^{-3}$, as discussed in \cite{kan_repo, kolmogrov_arnold_network_paper}.
After training/pruning, the model undergoes an automatic symbolification stage to derive an analytical representation of the learned mappings. The symbolification process utilizes a pre-defined set of functions, 
including the operations $\{\texttt{x}, \texttt{x}^2, \texttt{x}^3, \texttt{x}^4, \exp, \log, \sqrt{\phantom{x}}, \tanh, \sin, \tan, \texttt{abs}\}$ \cite{kan_repo}, as options
to approximate the KAN's spline-based components with closed-form expressions. The resulting analytical formula is extracted 
yielding an interpretable symbolic expression that closely approximates the trained KAN model. This enables transparent inspection of the nonlinear relationships learned by the network, thereby enhancing interpretability and providing insight into the underlying functional dependencies.
%


\subsection{ Results Discussion}

\subsubsection{Experiment~1: Training without Pruning}
\begin{figure}[t!]
\centering
\includegraphics[width=0.95\linewidth]{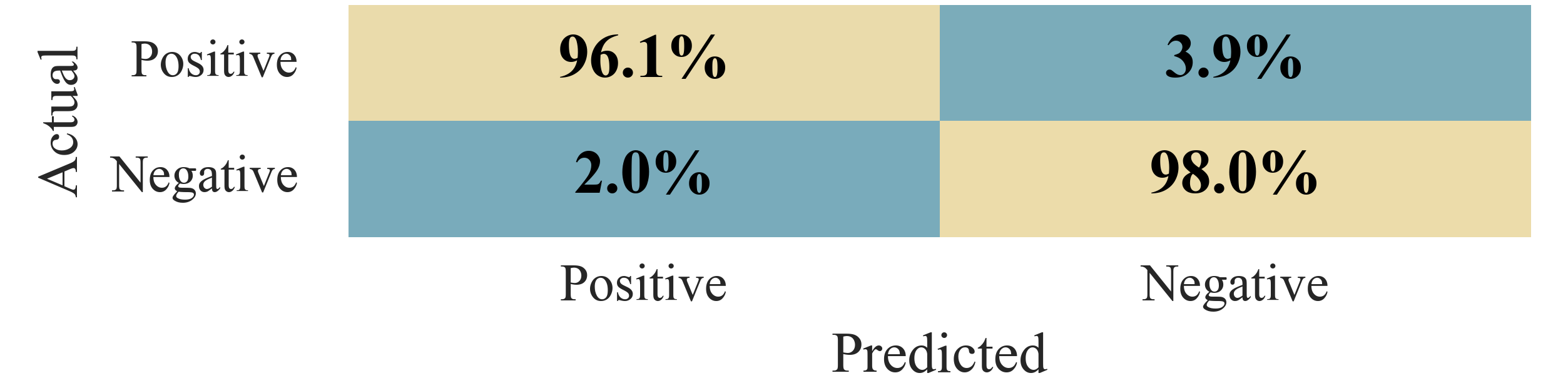}
\caption{Confusion matrix for Experiment~1 using KAN model $\mathcal{F}_{\theta}(\mathbf{x})$ (unpruned)}
\label{fig:KAN_CM}
\end{figure}

\begin{figure}[t!]
\centering
\includegraphics[width=0.95\linewidth]{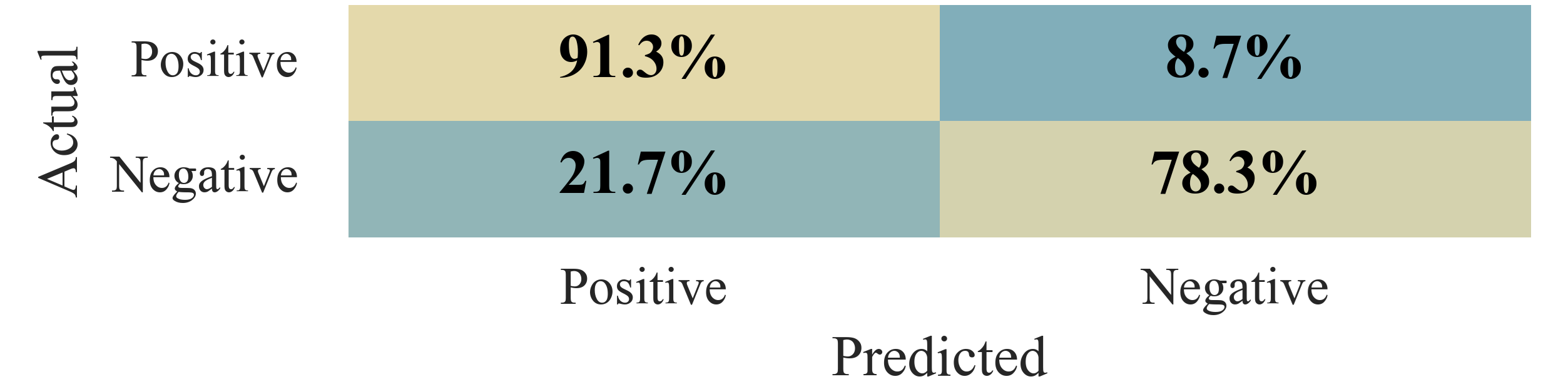}
\caption{Confusion matrix for Experiment~1 using extracted symbolic $\tilde{f}(\mathbf{x})$ (unpruned KAN)}
\label{fig:Symbolic_CM}
\end{figure}

The unpruned KAN achieves an accuracy of 97.28\%, precision of 98.02\%, recall of 96.50\%, and F1-score of 97.25\%. 
The confusion matrices in figures~\ref{fig:KAN_CM} and \ref{fig:Symbolic_CM} confirm
low false alarm rates and strong detection performance, indicating the model’s ability to capture the nonlinear dynamics of AGC behavior under both normal and attacked conditions.  
When the symbolic approximation, \(\tilde{f}(\mathbf{x})\), 
is extracted using the auto-symbolic process, its accuracy decreases to 83.55\%, representing a 13.7\% difference compared with the trained KAN. This degradation arises because the numerical B-spline activations within the KAN capture locally optimized nonlinear behaviors that may not perfectly map onto any single global symbolic function in the utilized set of functions supported by currently available KAN libraries. Nonetheless, the extracted expression still retains physically meaningful relationships among features, offering interpretive value even with reduced quantitative fidelity.

\subsubsection{Experiment~2: Model Pruning and Fine-Tuning}
\begin{figure}[t!]
\centering
\includegraphics[width=0.95\linewidth]{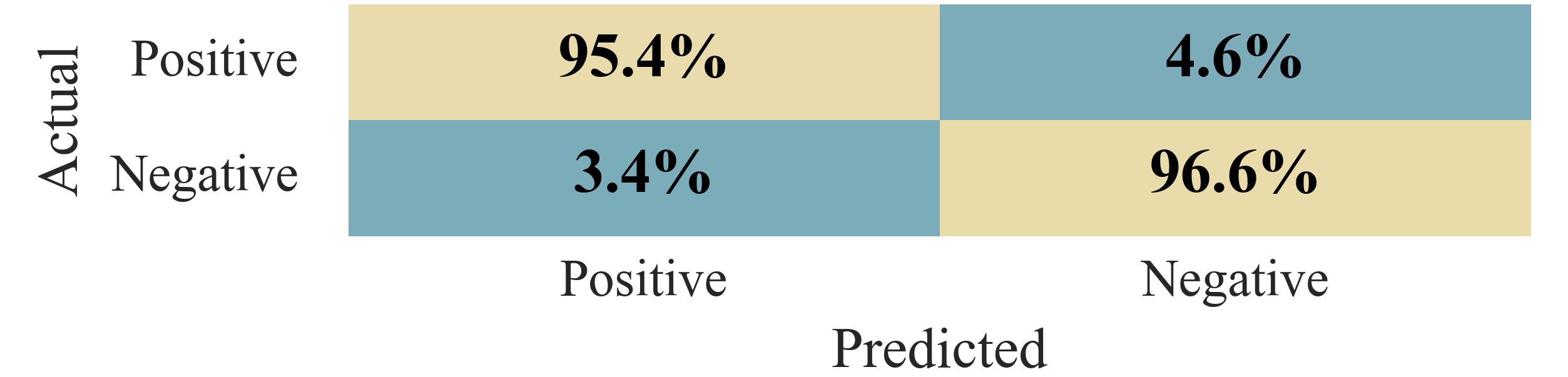}
\caption{Confusion matrix for Experiment~2 using KAN model $\mathcal{F}_{\theta}(\mathbf{x})$ (pruned and fine-tuned)}
\label{fig:KAN_CM_Exp2}
\end{figure}

\begin{figure}[t!]
\centering
\includegraphics[width=0.95\linewidth]{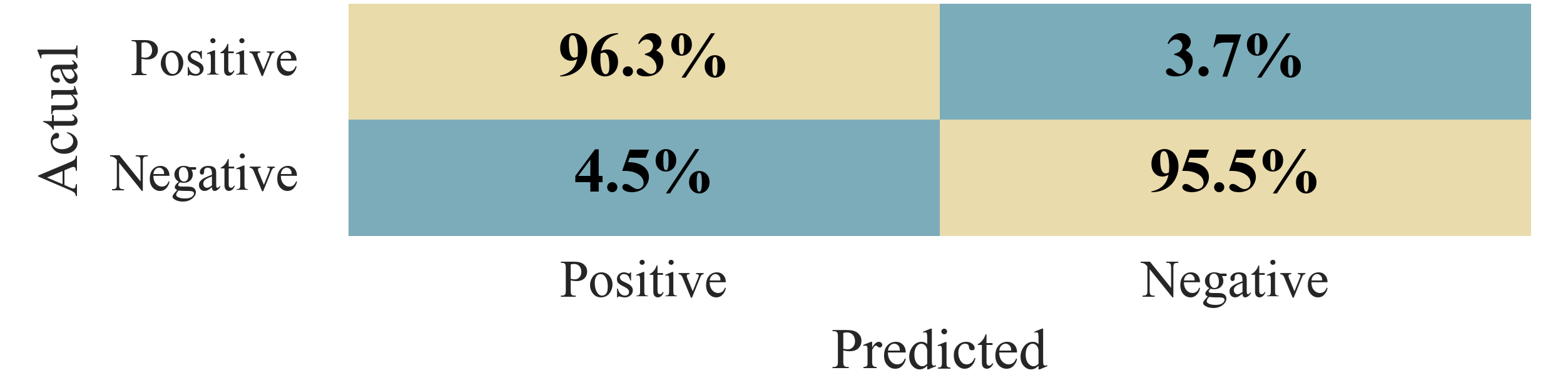}
\caption{Confusion matrix for Experiment~2 using extracted $\tilde{f}(\mathbf{x})$ (pruned and fine-tuned KAN)}
\label{fig:Symbolic_CM_Exp2}
\end{figure}

\begin{figure*}[t!]
\centering
\includegraphics[width=0.9\linewidth]{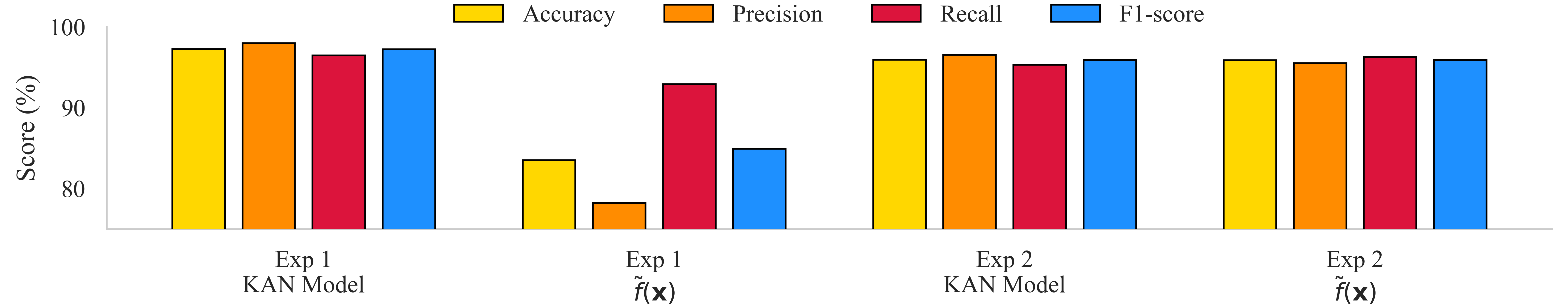}
\caption{Performance metrics for KAN model and symbolic expression \(\tilde{f}(\mathbf{x})\).}
\label{fig:KANresults}
\end{figure*}

To reconcile accuracy with interpretability, low-weight edges are pruned, followed by fine-tuning to restore predictive capacity. As depicted in Figures~\ref{fig:KAN_CM_Exp2} and \ref{fig:Symbolic_CM_Exp2}, the pruned and fine-tuned KAN achieves an accuracy of 95.97\%, precision of 96.56\%, and recall of 95.35\%. The corresponding symbolic expression \(\tilde{f}(\mathbf{x})\) reaches an almost identical accuracy of 95.9\%, with only a 0.07\% difference. This negligible discrepancy demonstrates that pruning and retraining yield a compact, interpretable model that preserves predictive power while producing an accurate symbolic representation.  
%
%
%
Fig.~\ref{fig:KANresults}
summarizes both experiments, confirming that pruning improves interpretability without significant accuracy loss. From a power system standpoint, this is vital: the model not only detects more FDIAs with minimal false alarms, but its interpretability improves operator confidence and overall AGC dependability.

%
%


%

\subsubsection{ Discussion}

Unlike post-hoc explainability tools such as SHAP or LIME, which provide feature importance estimates based on sample-specific perturbations, the proposed KAN framework delivers intrinsic interpretability.  
The symbolic expression is obtained directly from the trained model’s functional form before deployment, enabling engineers to verify and audit its decision logic without the need for surrogate explanations.  
This transparency enhances operational trust and facilitates integration of interpretable AI within critical control systems such as AGC.
While the complete symbolic expression offers full transparency, its mathematical complexity may limit immediate operator readability.  
For practical deployment, simplified or approximated forms of the expression, retaining only dominant terms, can provide human-understandable decision rules with minimal accuracy degradation.  
Developing such interpretable abstractions while preserving fidelity remains a promising direction for future work.

In detail, the extracted symbolic expression $\tilde{f}(\mathbf{x})$ obtained from the pruned KAN model 
provides a transparent mapping between the statistical features of AGC measurements and the model’s final decision boundary.  
Inspection of the dominant terms reveals that features associated with higher-order statistics of frequency measurements exhibit the largest coefficients, confirming their pivotal role in distinguishing manipulated measurements from measurements that correspond to genuine physical disturbances. This observation aligns with physical intuition, as FDIAs tend to disrupt the stochastic profile of frequency signals, increasing asymmetry and variability compared to legitimate events.  
Moreover, analysis of the functional components of $\tilde{f}(\mathbf{x})$ reveals the interplay between sinusoidal and polynomial terms. The sinusoidal components capture oscillatory dynamics characteristic of AGC response, while quadratic terms reflect nonlinear couplings between different AGC measurement statistics. 
This decomposition provides valuable interpretive depth: smaller signal magnitudes tend to emphasize constant or linear contributions, whereas large perturbations activate nonlinear terms, which is consistent with AGC behavior under stress or attack conditions.
It is worth noting that the symbolic formula extracted from the unpruned KAN model in \textit{Experiment~1} exhibits a much larger number of terms than that of pruned KAN, with a large number of small-magnitude coefficients and sinusoidal terms. 
This complexity indicates that several neurons in the unpruned model may lead to noisy functional components that degrade detection accuracy compared to the trained KAN, as discussed earlier. 
In contrast, the refined expression $\tilde{f}(\mathbf{x})$ obtained after model pruning and fine-tuning (Experiment~2) demonstrates both structural sparsity and higher functional coherence. 
This explains why $\tilde{f}(\mathbf{x})$ in \textit{Experiment 2} achieves near-identical performance to its parent KAN model, while $\tilde{f} (\mathbf{x})$ in \textit{Experiment 1} shows a more noticeable accuracy gap. 
Hence, symbolic extraction combined with post-pruning fine-tuning represents a key enabler for interpretable yet accurate FDIA detection in nonlinear AGC systems.

\subsection{Attack Detection Speed and Real-Time Feasibility}

In AGC systems, timely verification of measurement authenticity, by detection of possible FDIAs, is as critical. Even the most precise detector may lose operational value if its computational latency delays the AGC response or interferes with real-time control actions. Hence, it is essential to evaluate the inference speed of the proposed framework to ensure that it can identify malicious measurement manipulations promptly without hindering AGC functionality under normal operation.
In this experiment, the detection time is assessed using 4000 distinct samples representing a mix of FDIAs and normal disturbances. The inference time of the symbolic expression $\tilde{f}(\mathbf{x})$, extracted from the pruned and fine-tuned KAN model in Experiment~2, is evaluated on a workstation equipped with an 
Intel\textsuperscript{\textregistered} Core\textsuperscript{TM} Ultra 7 155H CPU operating at 1.4~GHz with 16~GB of RAM. 
Our results confirm that the proposed framework's average inference time is 0.153~seconds, i.e., per sample of FDIA.
This detection latency, of roughly 150~milliseconds per event, represents only a minor computational overhead compared to the timescale in which AGC systems operate, which is typically in the order of tens of seconds to minutes \cite{khalaf2018joint}.
Therefore, the proposed KAN's symbolic formula enables real-time detection of FDIAs within practical AGC time frames.


\subsection{Performance Comparison with Related Work}
\begin{figure*}[t!]
\centering
\includegraphics[width=0.9\linewidth]{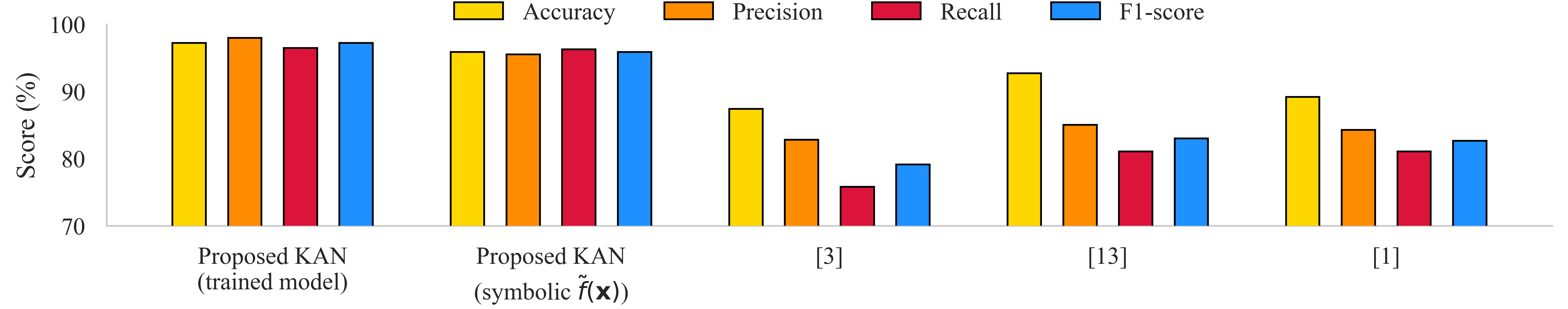}
\caption{Performance comparison between the proposed KAN and existing approaches}
\label{fig:KAN_vs_existing_approaches}
\end{figure*}

 To benchmark the performance of the proposed KAN-based framework, three previously proposed detection methods for AGCs, which mainly leverage LSTMs or LSTM-autoencoders directly on the raw AGC measurements, are re-implemented on our dataset generated above. 
Fig. \ref{fig:KAN_vs_existing_approaches}
summarizes the obtained results in this subsection.  
The proposed approach outperforms approaches proposed in related work, including deep learning methods such as LSTM autoencoders, achieving both higher accuracy and interpretability. Even the symbolic \(\tilde{f}(\mathbf{x})\) surpasses all black-box counterparts, demonstrating that transparency does not necessitate a trade-off with performance.  
Furthermore, the proposed framework requires only communicated AGC measurements, unlike some prior models that depend on unobservable internal states or plant-level data. This practical design ensures direct deployability in real AGC infrastructures without additional sensing overhead. The extracted symbolic representation also facilitates formal verification, model auditing, and operator understanding, capabilities absent from previous black-box approaches.  
Overall, the results establish that the KAN-based detection framework provides a realistic, interpretable, and high-performing layer of defense against FDIAs for nonlinear AGC systems, achieving accurate FDIA detection with minimal false alarms while maintaining operational trust and dependability. 


\section{Conclusion and Future Work}
\label{section:Conclusion}

This paper presented an interpretable learning-based framework for detecting FDIAs that target AGC systems by manipulating their measurements while explicitly accounting for inherent system nonlinearities such as governor deadband, generation rate constraints, and communication delays. The proposed method employs KANs to achieve both high detection accuracy and intrinsic interpretability, thereby bridging the gap between black-box learning models and transparent analytical formulations.
The proposed KAN-based approach was evaluated using a comprehensive dataset of FDIAs and physical disturbances with no FDIAs, using a nonlinear two-area AGC benchmark system. The trained KAN achieved a detection accuracy of 95.97\%, and its extracted symbolic counterpart, \(\tilde{f}(\mathbf{x})\), achieved nearly identical performance, 95.9\%, confirming that interpretability can be attained without compromising predictive reliability. 
The symbolic expression provides a direct analytical mapping between statistical features of communicated AGC measurements and detection outcomes, enabling operators to trace and validate the rationale behind each decision. 
%
Compared with existing deep learning–based methods such as LSTMs and autoencoders, the proposed framework achieves superior accuracy while maintaining transparency. 
Moreover, our results confirm the proposed method's capability of real-time attack detection.
%
%
The proposed framework  demonstrates that interpretable deep architectures like KANs can effectively reconcile model transparency with high  performance, an essential step toward trustworthy AI applications in critical infrastructure systems.
Future researchers can focus on optimizing KAN's mathematical expressions for more human interpretability and investigate the adoption of KANs for enhancing the cybersecurity of other critical smart grid components.

 \bibliographystyle{IEEEtran} 
 \bibliography{cas-refs.bib}


\end{document}